%% file: collas2022_conference.tex
\documentclass{article} 
\usepackage{collas2022_conference,times}

\input{math_commands.tex}

\usepackage{hyperref}
\hypersetup{
    colorlinks=true,
    linkcolor=red,
    filecolor=magenta,      
    urlcolor=blue,
    citecolor=purple,
    pdftitle={Overleaf Example},
    pdfpagemode=FullScreen,
    }

\usepackage{algorithm}
\usepackage{algpseudocode}
\usepackage{arydshln}
\newcommand{\stdv}[1]{\scriptsize$\pm$#1}
\usepackage{comment}
\usepackage{dsfont}
\usepackage{booktabs}
\usepackage{graphicx}

\usepackage{xcolor}

\usepackage{subfigure}
\usepackage{multirow}

\title{Open-Set Recognition via Augmentation-based Similarity Learning}

\author{Sepideh Esmaeilpour, Lei Shu, Bing Liu  \\
Department of Computer Science, 
University of Illinois at Chicago\\
\texttt{sesmae2@uic.edu, shulindt@gmail.com,  liub@uic.edu} \\
}

\collasfinalcopy 

\begin{document}

\maketitle

\begin{abstract}
The primary assumption of conventional supervised learning or classification is that the test samples are drawn from the same distribution as the training samples, which is called \textit{closed set learning or classification}. In many practical scenarios, this is not the case because there are unknowns or unseen class samples in the test data, which is called the \textit{open set scenario}, and the unknowns need to be detected. This problem is referred to as the \textit{open set recognition} problem and is important in safety-critical applications. We propose to detect unknowns (or unseen class samples) through learning pairwise similarities. The proposed method works in two steps. It first learns a closed set classifier using the seen classes that have appeared in training, and then learns how to compare seen classes with pseudo-unseen (automatically generated unseen class samples). The pseudo-unseen generation is carried out by performing distribution shifting augmentations on the seen or training samples. We call our method OPG (Open set recognition based on Pseudo unseen data Generation).
The experimental evaluation shows that the learned similarity-based features can successfully distinguish seen from unseen in benchmark datasets for open set recognition. 
\end{abstract}

\section{Introduction}\label{intro}
Deep classification models trained with cross-entropy loss are expected to produce a high probability score for the ground-truth class. Although this is the expected behaviour, the system can easily be fooled when encountering test instances from unknown or unseen classes. The reason is that in this case, the test data distribution does not follow the distribution of the train data. This situation cannot be avoided or ignored in open world settings. In fact, it is a well-studied phenomena that deep models are over-confident on unseen or distributionally shifted instances \citep{bendale2016towards}. Since the cross entropy loss aims at increasing the score for the ground-truth class in training, the learned features tend to be more discriminative than descriptive. Regardless whether it is from a seen or unseen class, as long as a test sample resides in a seen class's space (even if the test sample is far from the distribution of the seen class), it will be detected as a seen class sample. The detection of such out-of-distribution (OOD) samples is of critical importance for AI agents deployed in safety-critical applications such as medical diagnosis and autonomous driving because detecting unseen as normal/seen class samples can be catastrophic.

The terms \emph{open set recognition (OSR)} and \emph{out-of-distribution(OOD) detection} co-occur in the literature frequently. OSR methods often try to detect open-set samples by assuring that the closed-set accuracy remains intact. Early techniques for OSR are mainly SVM-based methods~\citep{scholkopf1999support, towards,scheirer2014probability}. Due to the great success of deep learning models in classification tasks, most recent methods for OSR are based on deep learning models. Particularly, since deep models tend to output high confidence scores for the majority of unseen or fooling samples, OSR cannot be directly performed based on the output scores. \citet{bendale2016towards} proposes a calibration method that uses the penultimate layer activation of a sample. Rejection or acceptance is done according to the calibrated OpenMax scores. Generative OpenMax \citep{ge2017generative} is an extension of OpenMax that follows the same protocol except that it uses generated samples to synthesize pseudo-unseen and trains a $k+1$ class classifier where class $k+1$th represents the space of unseen. \citet{perera2020generative} proposes a generative-discriminative technique that combines the advances in self-supervised learning and generative models. This method benefits from a generative data-augmentation technique per seen sample in training. A recent distance-based method \citep{miller2021class} encourages the seen class samples to form tight clusters around an anchored class center. The detection is performed based on the distance of the test sample in the logit space to the anchors.

On the other hand, out-of-distribution (OOD) detection techniques mainly concern with detecting OOD/novel samples rather than preserving the closed-set accuracy. Some of well-known OOD detection methods are based on input preprocessing techniques and softmax temperature scaling \citep{liang2017enhancing, hsu2020generalized}. In addition, various generative methods have been proposed for OOD detection \citep{ andrews2016detecting, chen2017outlier, perera2019ocgan}. Another recent set of methods for OOD detection exploits self-supervision techniques. These methods often learn a pretext task on the in-distribution data. Then OOD detection is performed based on the generalization error of this task on test samples \citep{bergman2020classification, hendrycks2019using}.

Recently, several authors distinguished \emph{far} and \emph{near} OOD detection problems\citep{winkens2020contrastive}. For instance, detecting OOD CIFAR100 from in-distribution CIFAR10 is considered a near-OOD (hard) problem as there are visually similar categories in these datasets. Detecting OOD CIFAR10
from in-distribution SVHN (photographed digits) is considered a far-OOD (easy) problem because their categories are visually and semantically very different. From this perspective, most open-set recognition techniques fall in the hard-OOD detection category since the closed-set and open-set classes used for evaluation contain visually similar categories in many cases.

Our focus in this work is OSR for image classification. We propose to actively learn similarity features for seen classes as well as for automatically generated pseudo-unseen classes in training. By doing so, in addition to seen classes, our model learns explicit pseudo-unseen classes in training. Then, the open set detection at test is done based on the similarity of a given test sample to seen and pseudo-unseen classes. 
The main contributions of our work (OPG) are as follows:

\emph{1)} We propose to detect open set samples by considering  both  the  similarity to seen samples as well as to pseudo-unseen samples. \emph{2)} Our pseudo-unseen sample generation is done by performing distribution shifting data augmentation on images of seen classes. Therefore, an advantage of the proposed method is that it does not require the training of an extra generative model along with the closed set classifier. \emph{3)} In our experiments, we demonstrate the effectiveness of the proposed method on benchmark datasets.
\section{Background}\label{background}
It is a common practice in image classification task to randomly augment training samples. Using various types of augmentations makes the model invariant to the visual differences of data points at test. Therefore, it improves generalization of the model on test data. Apart from this, specific types of augmentations such as rotations have shown to be one of the most effective ones for self-supervised training in computer vision. \citet{gidaris2018unsupervised} uses the prediction of rotation angle for a given image as a pretext task for self-supervised representation learning. This is mainly attributed to the fact that the trained model needs to attend to the actual objects in the image in order to predict the rotation angle correctly. Following this work, \citet{hendrycks2019using} proposes to improve model uncertainty and robustness against out-of-distribution samples by training using cross-entropy loss with an auxiliary rotation prediction loss. Auto-novel~\citep{han2020automatically} is a technique for novel class discovery via transfer learning. It  benefits from training of self-supervised RotNet~\citep{gidaris2018unsupervised} to learn general and transferable representations.

Furthermore, self-supervised and supervised contrastive learning frameworks~\citep{chen2020simple, supclr} owe their success to the careful composition of various augmentation techniques in generating positive and negative pairs of instances/samples for contrastive training. Particularly, \citet{chen2020simple} applies random cropping, random color distortion, and random Gaussian blur on an instance in order to generate an augmented version. Then, the instance and its augmented version is treated as a positive pair in contrastive loss. The authors demonstrate that the $90^\circ$, $180^\circ$, and $270^\circ$ rotations of the same image, when treated as a positive pair in contrastive loss, does not result in good representations on downstream tasks. Inspired by this observation, we hypothesize that 90x rotations of an image can simulate the distributional shift of unseen classes that the model encounters at test time. Therefore, we use pseudo-unseen data generated from rotations for feature similarity learning during training.

Our work is also related to meta classification loss (MCL), which was originally proposed by \citet{hsu2019multi} to do multi-class classification based on binary labels. Therefore, it learns a closed-set classifier. We integrate a sample generation and pairing strategy to the MCL framework to be able to conduct open set recognition. As we will explain, the proposed detection score of OPG is a practical consequence of our design choice. Our experimental results shows that rotated versions of seen images can simulate the actual unseen samples which only appear at test time.
\subsection{Problem Definition}
An open set recognition (OSR) problem is composed of two sub-tasks: 1) classifying the test samples from seen classes and 2) detecting samples that belong to unseen classes. Formally, the training data is represented by $D^S=\{(x_i^S,y_i^S)|i=1,\dots,N\}$, where $x_i^S$ is the training sample and $y_i^s \in \{1,...,r\}$ is its corresponding label. The number of seen classes is $r$. The test data is shown by $D^{test}=D^{S\cup U} = \{(x_i,y_i)|i=1,\dots,M\}$, which includes $M$ samples from both seen and unseen classes($S\cap U= \O$). The number of unseen classes is $q$ which is not known to the model at training or testing.
\section{Proposed Method}\label{method}
This section presents the proposed technique. We first motivate the use of a similarity-based binary cross-entropy loss for open set recognition. We then elaborate the generation process of pseudo-unseen samples and the pairing method that we apply. Finally, we explain how our two-step training works.
In an OSR problem, the detection scores are commonly designed based on the softmax scores (conditional class probabilities). It is natural to expect the distribution of two samples belonging to the same class to be consistent with each other. Therefore, we directly optimize a binary cross entropy loss in the probability outputs of the model. Some of the generative methods for OSR  define a dump/reject class to model the space of unseen. We argue that modeling the large space of unknowns in a single class is not the best choice as it does not account for similarities/differences among unseen classes. A model that can explicitly partition the unknown space is a better choice. However, this might sound infeasible as \textit{unseen} is meant to remain hidden during training. We consider using pseudo-unseen samples instead. The pseudo-unseen data can be external unseen data or the data generated from the seen classes. The former is restrictive as the external data has to be broad enough to represent possible unseen distributions and such data may not be available. 
Therefore, We build our method based on the latter which is the generation of pseudo-unseen through augmentations.

\subsection{Pseudo-Unseen Samples}

\begin{figure*}[t]
    \centering
    \includegraphics[width=0.9\linewidth]{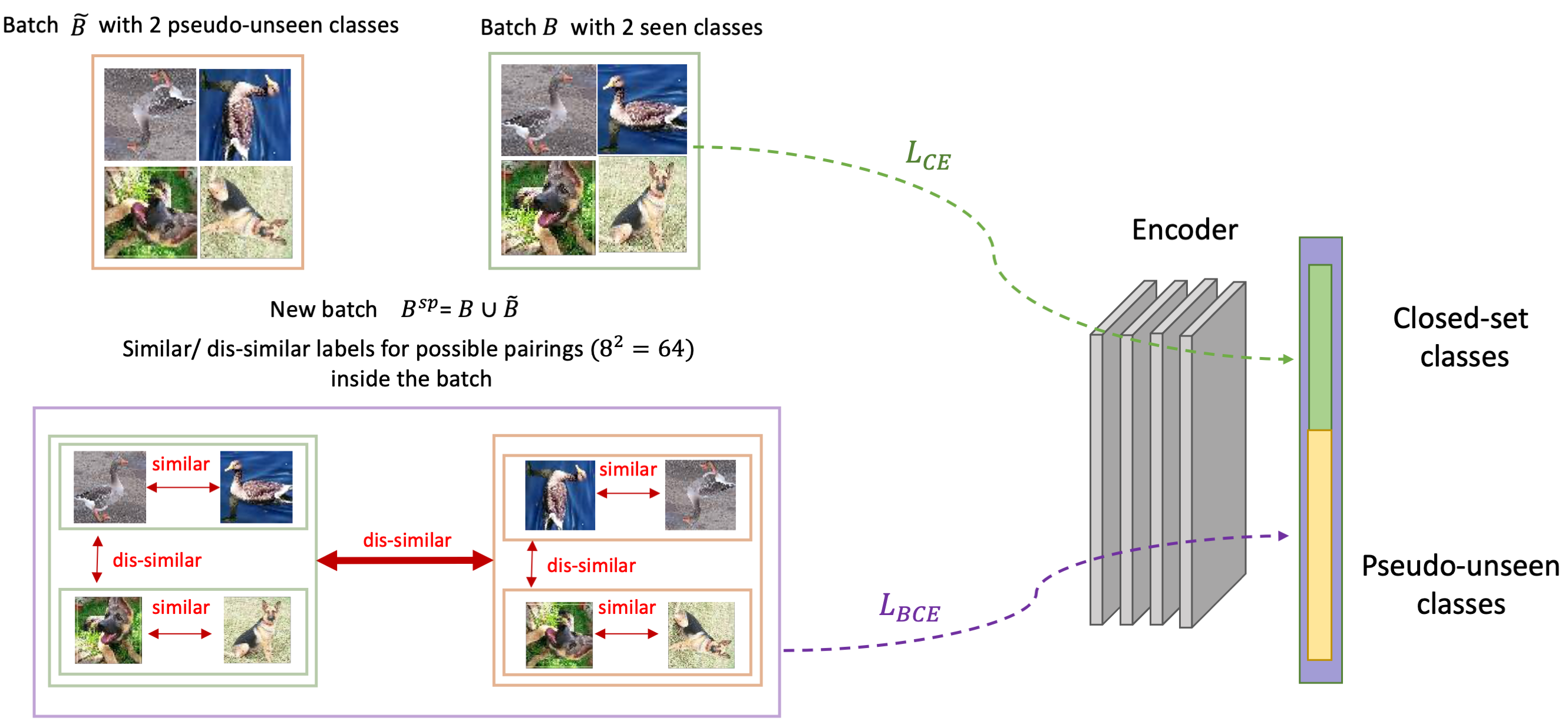}
    \caption{Illustration of pairing scheme based on the original images and their random rotations. Note that this is only a minimal example. A random batch of training samples with a sufficiently large batch size will possibly contain samples from all seen classes. The encoder in the right is trained on $r$ closed-set classes (green) in the first step of training. In the second step of training, the fully connected layer is extended to have $r+r^\prime$ nodes for seen (green) and pseudo-unseen (yellow) respectively. Best viewed in color.}\label{pairing_scheme}
\end{figure*}

Using geometric transformations, i.e. 90x rotations, has been studied in the context of self-supervised training \citep{gidaris2018unsupervised}. In addition, the experimental results of \citet{chen2020simple} show that two different rotations of the same image, when treated as a positive (similar) pair in the contrastive loss framework, do not result in good representations for the downstream tasks. In our case, we are interested in generating samples dissimilar to the original ones in the feature space. We hypothesize that the augmentations of the seen class samples, if chosen properly, can be regarded as pseudo-unseen data. This pseudo-unseen data can be used to expose the closed-set classifier to the actual unseen data that will appear in testing. The rotation is thus a good candidate. We have experimented with other augmentations like Gaussian blurring, Gaussian noise and color jitter, but rotation of images by multiples of $90^\circ$ achieves the best results.

Consider the seen class sample $(x_i^S, y_i^S)$ in batch $B$ of size $T$. We randomly rotate each $x_i$ by either $90^\circ$, $180^\circ$ or $270^\circ$. The newly generated samples are $(x_i^{PU}, y_i^{PU})$, where $PU$ stands for \emph{pseudo-unseen}. The size of the newly generated batch $\tilde{B}$ is also $T$. Note that rotations do not change the object class of a sample, i.e., an upside down "duck" is still a "duck". However, we give $y_i^{PU}$ a different label from the seen class label $y_i^S$ since we consider it to be a  \textit{distributionally shifted} version of the original sample. Then, a new batch of $B^{SP}=B \cup \tilde{B}$ is formed.
The ground-truth pairwise similarity between $(x_k, y_k) \in B^{SP}$ and $(x_l, y_l) \in B^{SP}$ is given by:
\begin{align}\label{pairwise_label}
     s_{kl}=
\begin{cases}
    1, &     y_k = y_l\\
    0, &     \text{otherwise}
\end{cases}
\end{align}
Figure.~\ref{pairing_scheme} gives an illustration of similar/dis-similar pairs. Batch $B$ (in green square) has a total of 4 samples from 2 seen classes. Batch $\tilde{B}$ (in yellow square) is the pseudo-unseen batch generated by random rotations of images in $B$. The union of these two batches including a total of 8 samples is shown at the bottom.  
Note that labels from different groups in Figure.~\ref{pairing_scheme} lead to a similarity score of $s_{kl}=0$.
\subsection{Training}
\textbf{Architecture:} We use a CNN encoder architecture as in many OSR baselines. This architecture is introduced in \citet{neal2018open}. It consists of 10 convolutional layers with filters of size $3\times3$ in each layer. Batch normalization and a leaky relu non-linearity is applied after each convolutional layer. The only difference is that we set the number of output nodes of the linear classifier layer equal to the number of seen+pseudo-unseen classes(green and yellow nodes shown in Figure.~\ref{pairing_scheme}). We denote the number of pseudo-unseen nodes with $r'$, 
which is a hyperparameter of our method that can be set to an arbitrary high value regardless of the actual number of unseen classes $q$. 
The logits output from the linear classifier layer is:
\begin{align}\label{encoder}
    z= \text{Classifier}(\text{Encoder}(x))
\end{align}
The length of z is $r+r'$. Therefore; the output probability distribution is over $r+r'$ classes. As we will explain in the next section, by doing so we will have a probability distribution on seens as well as pseudo-unseens. 
Algorithm.~\ref{algo} summarizes the training and testing phases for OPG. 

\textbf{Total Loss}
The total loss to be optimized is the weighted sum of cross-entropy loss and binary-cross entropy loss:
\begin{align}\label{jointloss}
      \mathcal{L}=\lambda_1 \mathcal{L}_{CE} + \lambda_2\mathcal{L}_{BCE}
\end{align}
In the joint loss objective, the cross-entropy loss is only applied to the samples in seen classes. We set $\lambda_2 = 0$ until the cross-entropy loss stabilizes. We call this the \emph{first step} of training. Then we set $\lambda_2 = 1$ and continue training on the total loss in the \emph{second step}. We need to continue training on cross-entropy loss in the second step to avoid forgetting. In fact, the two-step training plays an important role in the convergence of the proposed method. We conduct an ablation study in the experiment section to compare the two-step training with a cold-start joint model training (without the first step). 
$\mathcal{L}_{CE}$ is used to learn a closed set classifier on the first $r$ classes of the model.
\begin{align}\label{ce-loss}
    \mathcal{L}_{CE}= -\frac{1}{N}\sum_{i=1}^N 1_{c=y_i}\log(p(c|x_i)) \quad \text{ if } \quad x_i \in B
\end{align}

\textbf{Similarity-based Binary Cross Entropy Loss:}
The aim of $\mathcal{L}_{BCE}$ is to learn clusters 

for seen class samples as well as for pseudo-unseen samples by pairwise similarity supervision. We adopt the meta-classification loss \citep{hsu2019multi} for our purpose which is defined as follows:
\begin{align}\label{clus-loss}
    \mathcal{L}_{BCE}=
    -\frac{1}{M^2} &  \sum_{k=1}^M\sum_{l=1}^M[ s_{kl}log(p(x_k)^Tp(x_l))+ (1-s_{kl})log(1- p(x_k)^Tp(x_l))]
\end{align}
$M$ is the number of samples in the augmented training batch $B\cup \tilde{B}$. The probability distribution is over all classes, $r+r'$, for a distribution on both seen and pseudo-unseen classes.
This loss function uses the dot product of the probability distribution for samples $x_k$ and $x_l$ as the pairwise similarity measure. Consider the case where $x_k$ is from a seen class and $x_l$ is from a pseudo-unseen class. The ground-truth label $s_{kl}$ for this pair is $0$. Therefore, We expect the probability distributions $p(x_k)$ and $p(x_l)$ to be different. Since $p(x_k)$ and $p(x_l)$ are two normalized vectors, their dot product gives the cosine similarity which is minimized when $p(x_k)$ and $p(x_l)$ are perpendicular to each other. Similarly, for a positive pair $x_k$ and $x_l$, the loss is minimized when two probability vectors lie in the same direction.

Note that \citet{han2020automatically} exploited a similar form of loss to Eq.~(\ref{jointloss}). But they solve a different problem, identifying classes in the novel data via a transfer learning based clustering method. As a result, $\mathcal{L}_{BCE}$ in~\citet{han2020automatically} learns the similarity/dissimilarity only for samples of unseen classes, but in case of OPG, $\mathcal{L}_{BCE}$ learns similarities/dissimilarities between seen-seen, unseen-unseen and seen-unseen samples.

\begin{algorithm}[h]
	\caption{OPG training and detection}\label{algo}
	\begin{algorithmic}[1]
		\Require labeled seen data $D^s$, test data $D^{test}$
        \State \textbf{Training}
        \State Initialize \textit{Encoder} model with output layer of size $(r+r')$
		\State $ \ \ \ \ $ $\lambda_1 \gets 1$, $\lambda_2 \gets 0$
		\State $ \ \ \ \ $ train on $\mathcal{L} = \lambda_1\mathcal{L}_{CE}$ until convergence
		\State $ \ \ \ \ $ $\lambda_2 \gets 1$
		\Repeat 
				\State $ \ \ \ \ $  Get batch $B$ of size $|N|$ from $D^s$. 
				\State $ \ \ \ \ $ Apply one random rotation per sample in batch $B$ to generate batch $\tilde{B}$
				\State $ \ \ \ \ $ $B^{SP}=B \cup \tilde{B}$
				\State $ \ \ \ \ $  calculate $\mathcal{L}_{BCE}$ for all $(x_i, x_j) \in B^{SP} \times B^{SP}$
				\State $ \ \ \ \ $ calculate $\mathcal{L}_{CE}$ only on $B$
				\State $ \ \ \ \ $ Backpropagate $\mathcal{L} = \lambda_1\mathcal{L}_{CE}+\lambda_2\mathcal{L}_{BCE}$
		\Until{convergence}
		\State \textbf{Testing(detection)}
		\For{$x^{test} \in D^{test}$} 
		\State $ \ \ \ \ $ Calculate $S(x_i^{test})$ according to Eq.~\ref{rej_criterion}
		\EndFor

	\end{algorithmic}
\end{algorithm}

\subsection{Testing}
Our model learns the similarity between seen and pseudo-unseen classes during training. In testing, when a sample is more similar to a pseudo-unseen class, the similarity-based  features will pull it towards one of the pre-allocated pseudo-unseen classes and push it away from the seen class centers. Therefore, the accumulative probability scores over pseudo-unseen classes can naturally be inferred as the confidence of the model in detecting a test sample as unseen.
 As the result, the open set detection score is defined as:
\begin{align}\label{rej_criterion}
S(x) = 1 - \sum_{\substack{0 \leq c< r}} p(c|x)
\end{align}
where $p(c|x)$ is the softmax output for seen class $c$. This score is used to calculate AUROC (Area under the ROC curve for open set detection). 
This section evaluates the proposed method, present the experimental setting, and analyze the experimental results.

\subsection{Datasets}
 The difficulty level of an OSR task is measured by the openness metric defined in \citet{scheirer2012toward}. A task is harder when more unseen classes are presented to the model at the test time. Openness is defined as  $Openness = (1-\sqrt{\frac{2*N_{train}}{N_{test}+N_{target}}})*100$ 
Where $N_{train}$ is the number of seen classes, $N_{target}$ is the number of seen classes at testing and $N_{test}$ is the total number of seen and unseen classes at test. Following the protocol used in recent OSR techniques, we evaluate the performance of our proposed method using the following datasets.

\emph{CIFAR10.}\citep{krizhevsky2009learning} The training set of seen classes is a set of 6 classes of CIFAR10. The 4 remaining classes are used as open set (unseen) classes.  (Openness = $13.39\%$)
\emph{CIFAR+10.}\citep{krizhevsky2009learning} This dataset uses 4 non-animal classes of CIFAR10 as the closed set (seen) classes for training. 10 animal classes are randomly chosen from CIFAR100 as the open set (unseen) classes. (Openness = $33.33\%$)
\emph{CIFAR+50.}\citep{krizhevsky2009learning} This dataset uses 4 non-animal classes from CIFAR10 as a closed set. All 50 animal classes from CIFAR100 are used as the open set classes. So it is a harder task than CIFAR+10. (Openness = $62.86\%$)
\emph{SVHN.}\citep{netzer2011reading} 6 random classes from SVHN (street view house numbers) are used for closed set training while the remaining 4 classes are used as open set (unseen) classes. (Openness = $13.39\%$)\\
\emph{TinyImagenet.}\citep{le2015tiny} It is a 200-class subset of ImageNet. 20 random classes are used as the closed set (seen) classes. The remaining 180 classes are used as open set (unseen) classes. (Openness = $57.35\%$)

The reported scores are averaged over 5 seen-unseen split for each dataset. The class splits that we have used are publicly available in the github repository of \cite{miller2021class} \footnote{https://github.com/dimitymiller/cac-openset}
\subsection{Baselines}
We compare OPG with 8 OSR baselines. \emph{DOC}~\citep{shu2017doc} is a method originally proposed for open set recognition of text data. It uses one-vs-rest sigmoid function in the output layer. It compares the maximum score over sigmoid outputs to a predefined threshold to reject or accept a test sample.
\emph{OpenMax}~\citep{bendale2016towards} is an early technique for open set recognition in deep models. It does calibration on the penultimate layer of the network to bound the open space risk.
\emph{G-OpenMax} and \emph{OSRCI} \cite{ge2017generative}\citep{neal2018open} are both generative models that use the set of generated samples to learn an extra class. So, the model is a $K+1$ class classifier of seen and pseudo-unseen.
\emph{C2AE} \citep{oza2019c2ae} is a class-conditioned generation method that uses the reconstruction error of unseen samples as the detection score.
\emph{CAC}\citep{miller2021class} is a recent approach that uses anchored class centers in the logit space to encourage forming of dense clusters around each known class. Detection is done based on the distance of the test sample to these seen class centers.
\emph{GFROR} \citep{perera2020generative} combines generation-based models with self-supervision learning to solve the problem.
\emph{Generalized ODIN} \citep{hsu2020generalized}
Generalized ODIN (G-ODIN) is a recent out-of-distribution (OOD) detection method which uses a decomposed confidence score for OOD detection. However, the code of G-ODIN have not been released. We thus could not run their official code for comparison. We implemented the G-ODIN algorithm by closely following the algorithm in their paper, which produced the results reported in Table~\ref{auc_table}. We also applied the same hyperparameters suggested in their paper. Table~\ref{auc_table} shows that G-ODIN does not do well for the OSR setting, which was called \textit{semantic shift detection} in \citet{hsu2020generalized} and the authors explicitly stated that semantic shift detection is a more challenging problem for G-ODIN than OOD detection.
ARPL \citep{chen2021adversarial} is a recent baseline. It proposes a classification framework with the adversarial
margin constraint to reduce the open space risk. In order to be consistent with our results, we generated the reported scores in Table~ \ref{auc_table} by running ARPL official code on the same seen and unseen splits as OPG.
\subsection{Evaluation Metrics}
\textit{AUROC}: Area under the ROC Curve is the primary measure for open set detection performance. It is threshold-free and indicates the trade-off between true positive rate (seen samples correctly detected as seen) and the false positive
rate (unseen samples incorrectly detected as seen). \textit{Classification accuracy:}
Classification accuracy is used to show the performance of our method when only seen classes (closed-set samples) are input to the model. 
\subsection{Implementation Details}
We use the same encoder architecture as the baselines \citep{neal2018open, perera2020generative, oza2019c2ae, miller2021class} for a fair comparison.
The encoder is a CNN architecture with a total of 10 convolutional layers. Batch normalization and leaky relu (0.2) are applied after each convolutional layer. For DOC \citep{shu2017doc}, we conducted the experiments with the same encoder with a sigmoid layer in the output as required by DOC algorithm.  Since we use 90x rotations to achieve distribution shift from the original data points, we do not apply random rotations on the original set of data points as part of the standard augmentations for classification problems.
The size of the output layer of the encoder is set to $r+r^\prime $(\textit{seen+pseudo-unseen}). $r^\prime$ is a hyperparameter of OPG, which we will discuss in Section~\ref{unknown_num}. We set the total number of classees to $r+r^\prime=100$ for CIFAR10, CIFAR+10, CIFAR+50 and SVHN experiments. For Tinyimagenet we set $r+r^\prime=200$. We use Stochastic Gradient Descent (SGD) to optimize the objective. 
At the first step of training ($\lambda_1 = 1$, $\lambda_2 = 0$) we train the model for 200 epochs with a batch size of 128 for CIFAR10, CIFAR+10, CIFAR+50 and SVHN. The initial learning rate is $0.01$ and it is decayed by a factor of 10 at 100th and 150th epochs respectively. For TinyImagenet, we train the model for 500 epochs with a batch size of 128. The initial learning rate is $0.01$ and it is decayed by a factor of 10 at the 300th epoch.
 In the second step, when we optimize  both $\mathcal{L}_{CE}$ and $\mathcal{L}_{BCE}$ ($\lambda_1 = 1$, $\lambda_2 = 1$) the network is trained with the batch-size of 256 for 100 epochs and a constant learning rate of 0.01 is maintained for CIFAR10, CIFAR+10, CIFAR+50 and SVHN. 
For Tinyimagenet, we train for 200 epochs with a batch size of 128. The learning rate is 0.01 and it is decayed by a factor of 10 at 100th epoch. 
\subsection{Results and Discussions}
\begin{table*}[h]
     \caption{Open set detection performance in terms of AUROC (\%). The results are averaged over 5 random splits of each dataset ($\pm$ standard deviation). All scores except the results of DOC and G-ODIN are taken from \cite{miller2021class}. The best scores are in bold. For TinyImagenet, the AUROC score of OPG is comparable to the best score. The last column gives the average value of each row or system.
     }\label{auc_table}
    \begin{center}   
    \resizebox{\textwidth}{!}{
    \begin{tabular}{l|cccccc}
    \hline
        &CIFAR10&CIFAR+10&CIFAR+50&SVHN&TinyImageNet&Average\\ 
    \hline
    OpenMax \cite{bendale2016towards} & 69.5\stdv4.4&81.7 \stdv NR& 79.6\stdv NR &89.4\stdv1.3 & 57.6\stdv NR&75.6\\
    G-OpenMax \cite{ge2017generative} & 67.5\stdv 4.4 & 82.7 \stdv NR & 81.9 \stdv NR & 89.6 \stdv 1.7 & 58.0 \stdv NR&75.9 \\
    DOC \cite{shu2017doc} & 66.5\stdv6.0 & 46.1\stdv1.7 & 53.6\stdv0.0 &74.5\stdv3.2 & 50.2\stdv0.5&58.2 \\
    OSRCI \cite{neal2018open} & 69.9\stdv3.8 & 83.8\stdv NR & 82.7\stdv 0.0 & 91.0\stdv 0.1 & 58.6\stdv NR&77.2 \\
   C2AE \cite{oza2019c2ae}
   & 71.1\stdv 0.8 & 81.0\stdv0.5 & 80.3\stdv0.0 &89.2\stdv1.3 & 58.1\stdv1.9&75.9 \\
    GFROR \cite{perera2020generative} & 80.7\stdv 3.0 & 92.8\stdv0.2 & 92.6\stdv0.0 &93.5\stdv1.8& 60.8\stdv1.7&84.0 \\
    G-ODIN \cite{hsu2020generalized}& 72.2\stdv 9.6 &51.7\stdv 0.9  &89.8\stdv0.0 &84.1\stdv 5.7 & 57.0\stdv 0.2&70.9 \\    
    CAC \cite{miller2021class} & 80.1\stdv3.0 & 87.7\stdv1.2 & 87.0\stdv0.0 &94.1\stdv0.7 & \textbf{76.0\stdv1.5}&84.9 \\
        ARPL\cite{chen2021adversarial}& 83.1\stdv 4.1 & 91.8\stdv0.7 & 92.3\stdv0.0 & \textbf{94.2\stdv 0.5} & 74.0\stdv3.0&87.0\\
        \hdashline
        OPG\textsuperscript{*}& 82.4\stdv2.6 & 89.6 \stdv0.3  & 90.4\stdv0.0&81.2\stdv1.7 & 73.5\stdv2.0 & 83.4 \\ 

        \textbf{OPG(ours)}& \textbf{83.1\stdv5.0} & \textbf{96.2 \stdv0.5 } & \textbf{96.1\stdv0.0}&89.0\stdv2.8 & 75.6\stdv2.6&\textbf{88.0}\\  
    \hline        
    \end{tabular}}
    \end{center}
\end{table*}
The experimental results of our proposed OPG and 8 baselines are summarized in Table~\ref{auc_table}. The last column gives the average score of each system or row on all datasets. We can see that on average OPG outperforms all baselines. Considering ARPL as the second best performing method, OPG performs better than ARPL on CIFAR10, CIFAR+10, CIFAR+50, and Tinyimagenet. Based on these results, we can conclude that rotations can successfully shift the distribution of seen classes in the case of natural images such as CIFAR10 and TinyImagenet and help produce effective OSR models. On the other hand, the performance on SVHN indicates that rotations of seen samples cannot simulate the actual unseen class samples. We suspect that the rotations might have a misleading effect in generating pseudo-unseen samples for digit datasets. For example, some images rotated by $\{90^\circ, 180^\circ, 270^\circ\}$ can still be very similar to the original ones(e.g., $8$, $0$ or $5$ still look similar to the original after a $180^{\circ}$ rotation). We can conclude that such noisy pairwise labels negatively affect the open set detection performance on digit datasets.
\begin{figure*}[h]
\centering
\subfigure[The effect of used augmentations]{%
{\includegraphics[width=0.45\linewidth, height=0.28\textwidth]{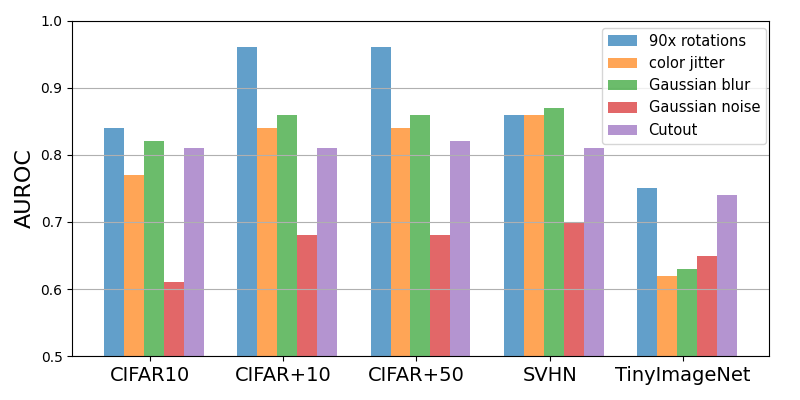}%
\label{augmentation_bars}%
}}\qquad
\subfigure[TSNE visualization]{%
\includegraphics[width=0.4\linewidth, height=0.3\textwidth]{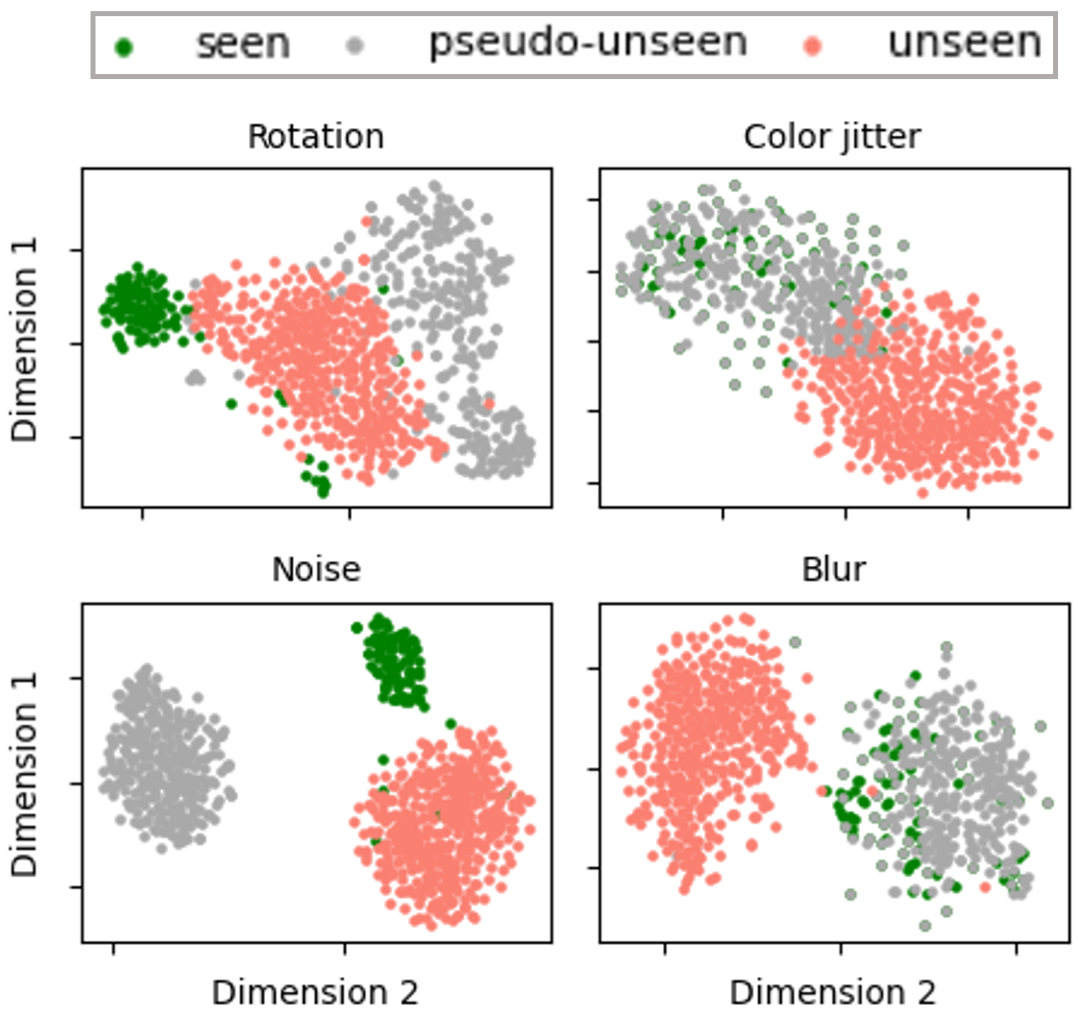}%
\label{4tsne}%
}
\caption{(a) The effect of using 5 types of augmentations applied to seen samples to generate pseudo-unseen data for CIFAR10. (b)TSNE visualization of seen, pseudo-unseen and unseen samples where pseudo-unseen is generated by seen+rotation, seen+color jitter, seen+noise and seen+blur in each subplot. (Best viewed in color)}
\end{figure*}
Figure~\ref{augmentation_bars} shows the AUROC scores when OPG is trained by using 5 different types of augmentations to generate pseudo-unseen samples. According to the results, OPG works best when 90x rotations are used for pseudo-unseen generation. We argue that using other types of augmentations including Gaussian blur, noise, color jitter or cutout can merely generate easy-to-detect pseudo-unseen data. In this case, the model can easily learn to discriminate these low-level features during training without focusing on high-level semantic features. On the other hand, 90x rotation of an image can generate hard-to-detect pseudo-unseen samples. The results summarized as OPG\textsuperscript{*} in Table \ref{auc_table} strengthens our claim. We used random rotations between $30^\circ$ to $60^\circ$ to generate pseudo-unseen samples for OPG\textsuperscript{*}. The detection performance of OPG\textsuperscript{*} degrades compared to OPG with 90x rotation augmentations. As noted by \cite{gidaris2018unsupervised} rotating images by multiples of $90^\circ$ do not cause easy-to-detect artifacts on the images which in our case leads to the generation of hard-to-detect pseudo-unseen samples. On the other hand, rotations in range of $30^\circ$ to $60^\circ$ generate easy-to-detect pseudo-unseen samples, hurting the open-set detection at inference.

It is worth noting that many contrastive training methods rely on careful composition of various augmentation types. However, their effectiveness is only shown empirically and there is little theoretical investigation of why a specific setting or composition yields better results than others.
 Figure~\ref{4tsne} demonstrates the differences of learned features when each of these augmentations used for the pseudo-unseen generation on CIFAR10.  The top-left plot shows seen, rotated seen and unseen classes in green, gray and orange respectively. The rotated seen (pseudo-unseen) and actual unseen classes form a mixed cluster of orange and gray while the seen class forms a separate cluster. This confirms our expected behaviour which is 1) the similarity of actual unseen to pseudo-unseen and 2) the dis-similarity of unseen to seen samples in the feature space. In the other three plots, either condition 1 or 2 are not satisfied. 
\subsubsection{Seen and Unseen clusters in the Feature Space}
\begin{figure*}[h]
\centering
\subfigure[TSNE visualization]{%
{\includegraphics[width=0.45\linewidth, height=0.28\textwidth]{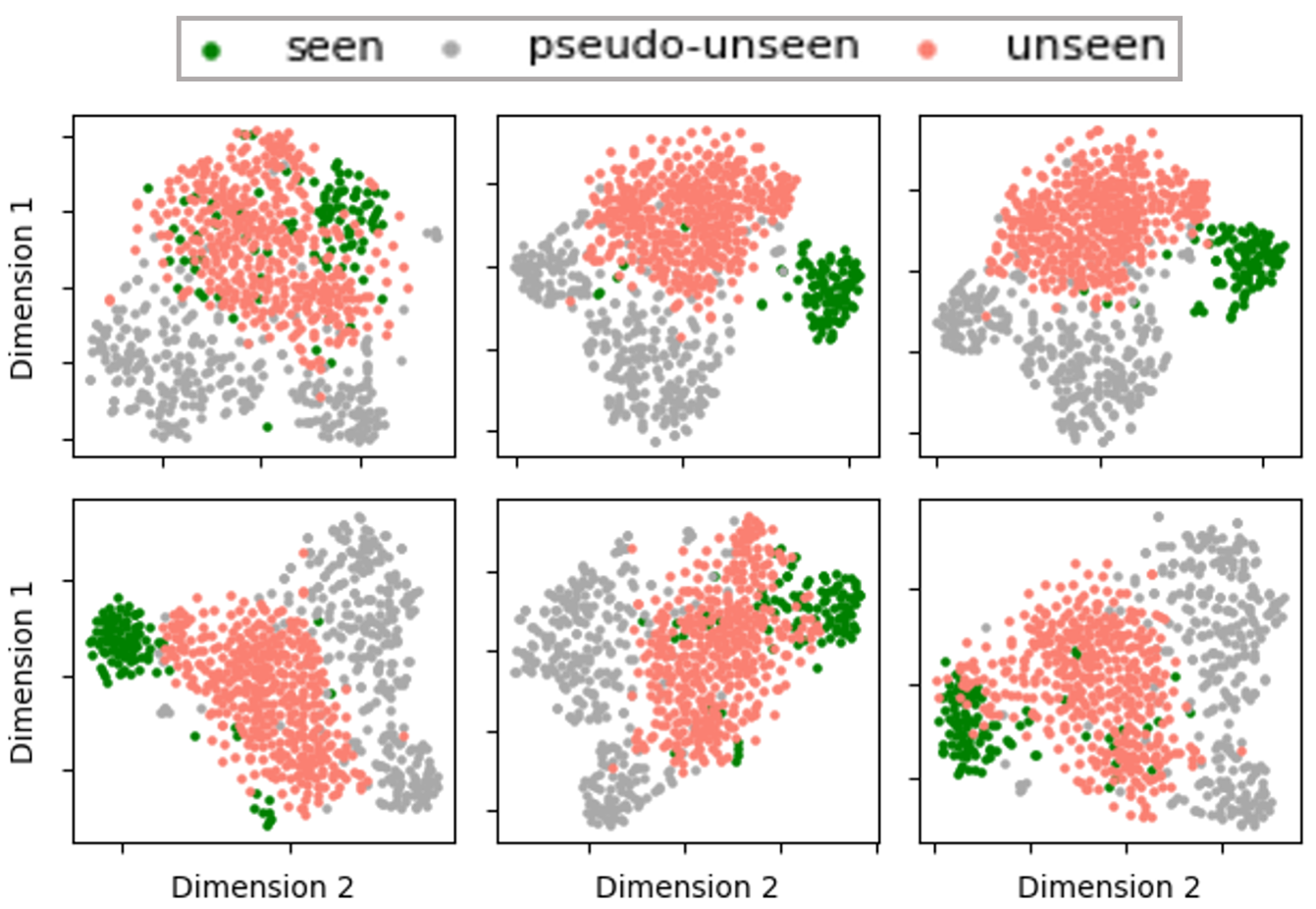}%
\label{6tsne}%
}}\qquad
\subfigure[Confidence score histogram]{%
\includegraphics[width=0.3\linewidth, height=0.25\textwidth]{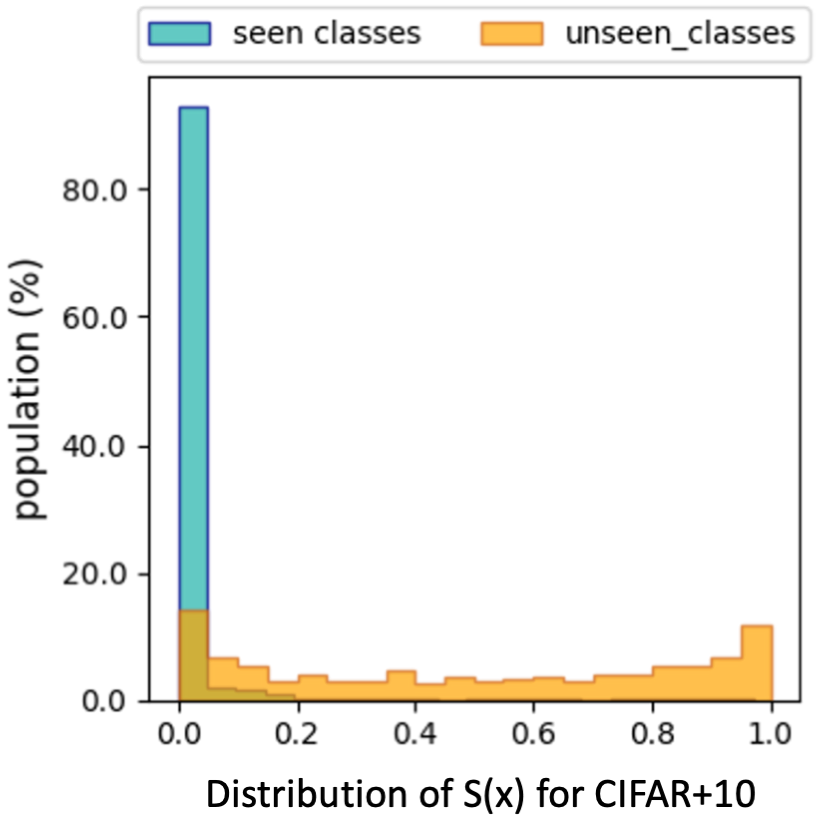}%
\label{histograms}%
}
\caption{(a) Each subplot shows a TSNE visualization of a seen class from CIFAR10 in green, the features of the rotated versions of the same seen class in gray and the actual 4 unseen classes in orange. (Best viewed in color) (b) The distribution of detection score $S(x)$ over seen and unseen classes of CIFAR+10, and SVHN respectively. (Best viewed in color)}
\end{figure*}

 We conducted a case study on CIFAR10 to investigate the learned feature space by OPG. Figure.~\ref{6tsne} illustrates where the actual unseen classes reside in the feature space compared to the seen classes and pseudo-unseen classes (the  $90^\circ, 180^\circ, 270^\circ$ rotations of the same seen samples). Ideally, our technique should learn a feature space where pseudo-unseen features are a simulation of actual unseen classes. TSNE \citep{vanDerMaaten2008} visualization shows that in most cases, unseen classes (in orange) and pseudo-unseen classes (in gray) can be interpreted as one large cluster compared to the seen classes which form separate clusters. This confirms that the distribution shift by rotations can simulate the actual shift from seen classes to unseen classes in testing. The unseen class samples at test time are closer to the generated pseudo-unseen samples than samples that the model has seen in training. Therefore, it can successfully distinguish such samples in testing. Figure.~\ref{histograms} shows the distribution of the confidence score $S(x)$ on a seen and unseen class of CIFAR+10.
\subsubsection{Unknown Number of Unseen Classes}\label{unknown_num}
Since OPG pre-allocates the pseudo-unseen classes in training, it is natural to think that by knowing the actual number of unseen classes we can learn a better open set classifier. To study the effect of the total number of training classes $r+r^\prime$ on OSR  performance, we performed an ablation study. Table.~\ref{numclus_table} illustrates how the AUROC score is affected when the total number of classes in training is varied from the ground-truth number of seen+unseen classes to a large number of classes.
\begin{table}[H]
    \caption{The effect of the number of pre-allocated pseudo-unseen classes is summarized in the table.
    We can confirm that knowing the ground truth $r+r'$ apriori does not affect the performance of OPG.}\label{numclus_table} 
    \begin{center}
    \resizebox{0.5\textwidth}{0.08\textwidth}{
    \begin{tabular}{cc}     
             \multicolumn{2}{c}{CIFAR10 / SVHN / CIFAR+10 / CIFAR+50 / TinyImageNet}\\
             \toprule

             $\mathbf{r+r^\prime}$& AUROC\\
            \hline    
             
              \textbf{\{10/10/14/50/200\}}  & 0.83 / 0.89  / 0.96 / 0.96 / 0.75\\
              \textbf{\{50/50/50/100/250\}} & 0.82 / 0.88 / 0.96 / 0.96 / 0.76\\
              \textbf{\{100/100/100/150/300\} }& 0.83 / 0.89 / 0.96 / 0.95 / 0.75\\ 
              \textbf{\{200/200/200/200/350\} } & 0.83 / 0.88 / 0.95 / 0.96 / 0.74\\
    \hline
    \end{tabular} 
    } 
    \end{center}
\end{table}
The first row corresponds to the case $r+r'=seen+unseen$. For the other 3 experiments, $r+r'>seen+unseen$. 
We can confirm from the table that the number of pre-allocated $r+r^\prime$ is not a major factor in OSR performance. Consider the experiment with the CIFAR10 dataset. The ground truth number of unseen classes is 4. In the extreme case experiment, we have set this number to $194(200-6)$ in training. Intuitively, this situation can be interpreted as having many empty pseudo-unseen clusters at training. Since the probability values for the extra pre-allocated classes $r\prime$ is very low, it does not affect the dot product value in $\mathcal{L}_{BCE}$; therefore, the performance remains stable.

\subsubsection{Two Step Training}
\begin{figure*}[h]
\centering
\subfigure[Two step training effect]{%
{\includegraphics[width=0.45\textwidth, height=0.2\textwidth]{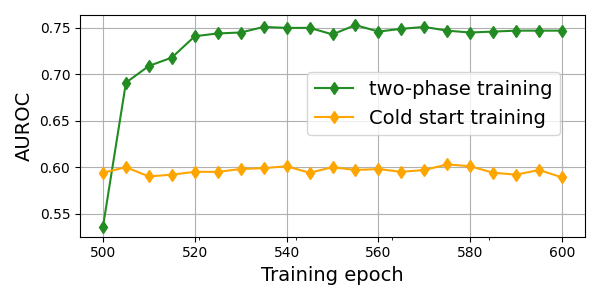}%
\label{coldstart}%
}}\qquad
\subfigure[closed-set accuracy after each training step]{%
\includegraphics[width=0.4\textwidth]{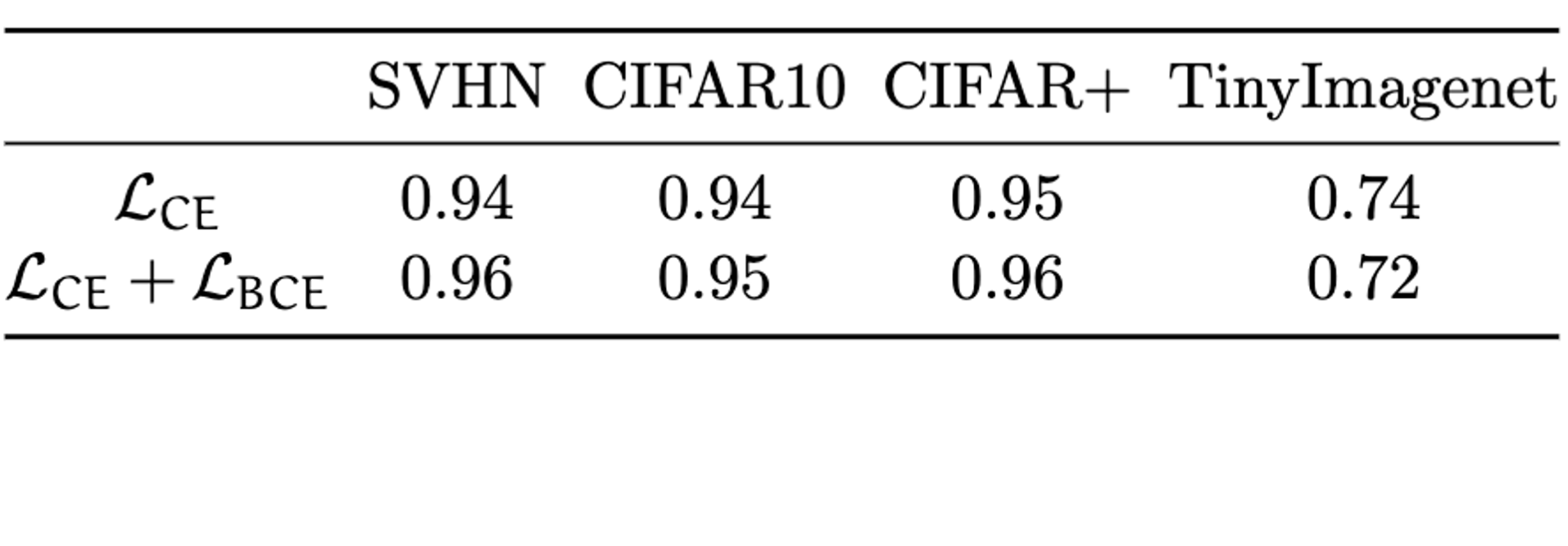}%
\label{accuracy}%
}
\caption{(a) The AUROC scores on test data for the last 100 epochs of training on TinyImagenet. Note that cold-start training optimizes the total loss for 600 epochs. In two-step training, the training starts by optimizing $\mathcal{L}_{CE}$ for 500 epochs, and then the joint loss is optimized for the last 100 epochs of training. Best viewed in color. (b) Closed-set accuracy recorded at the end of first and second step of training.}
\end{figure*}

Our proposed total loss is composed of two loss terms $\mathcal{L}_{BCE}$ and $\mathcal{L}_{CE}$. It might seem sufficient to optimize on the joint loss $\mathcal{L}_{CE}+\mathcal{L}_{BCE}$ together in the first place. However, $\mathcal{L}_{BCE}$ requires stable probability outputs of the network for similarity calculation. By training the encoder in a cold-start fashion, the model might converge to a local minimum. We did an ablation experiment on TinyImagenet to study the importance of the first warm-up step. Figure~\ref{coldstart} illustrates the AUROC score evolution on the test data as the training reaches the final 100 epochs. It is clear from the plot that the cold-start training has difficulty in convergence. Moreover, we observed that the closed set accuracy on seen classes tends to improve after the second step of training finishes. We have tabulated the classification accuracy at the end of the first step and second step of training. According to the Table~\ref{accuracy}, $\mathcal{L}_{BCE}$ not only contributes to similarity learning for OSR but also improves the accuracy on almost all datasets. TinyImagenet accuracy drops by only 2\%. This is intuitive as $\mathcal{L}_{BCE}$ groups samples of each seen class together far from other seen and pseudo-unseen clusters.

\section{Conclusion}
We proposed an open set recognition method that models pseudo-unseen classes along with seen classes at training. Our technique benefits from distribution shifting data augmentations to automatically generate pseudo-unseen samples; therefore, it eliminates the need for training generative models. Our method pre-allocates pseudo-unseen classes at training. A practical consequence of this design choice is a natural detection score that is simply the summation over softmax probabilities of pseudo-unseen classes at test. We showed in our experiments that the open set detection performance is independent of the number of pseudo-unseen classes that we use at training which makes our approach flexible in real world settings. 
The proposed method can perform better or similar to other state of the art baselines on benchmark datasets for open set recognition.

\section*{Acknowledgments}
{\color{black}This work was supported in part by two National Science Foundation (NSF) grants (IIS-1910424 and IIS-1838770), a DARPA contract HR001120C0023, and a Northrop Grumman research gift.} 

\bibliography{collas2022_conference}
\bibliographystyle{collas2022_conference}

\end{document}

%% file: math_commands.tex
\usepackage{amsmath,amsfonts,bm}

\def\eqref#1{equation~\ref{#1}}

\def\1{\bm{1}}

\DeclareMathAlphabet{\mathsfit}{\encodingdefault}{\sfdefault}{m}{sl}
\SetMathAlphabet{\mathsfit}{bold}{\encodingdefault}{\sfdefault}{bx}{n}

